\documentclass[conference]{IEEEtran}
\IEEEoverridecommandlockouts
\usepackage{cite}
\usepackage{amsmath,amssymb,amsfonts}
\usepackage{algorithm}
\usepackage{algpseudocode}
\usepackage{subfig}
\usepackage{graphicx}
\usepackage{textcomp}
\usepackage[table]{xcolor}
\usepackage{siunitx}
\usepackage{pdfpages}
\usepackage{rotating}
\usepackage{lscape}
\usepackage{footmisc}
\usepackage{caption}
\newcommand\R{{\mathbb R}}

\def\tablename{Table}
\usepackage{etoolbox}
\makeatletter
\patchcmd{\@makecaption}
  {\scshape}
  {}
  {}
  {}
\patchcmd{\@makecaption}
{\\}
{:\ }
{}
{}
\makeatother

\def\BibTeX{{\rm B\kern-.05em{\sc i\kern-.025em b}\kern-.08em
    T\kern-.1667em\lower.7ex\hbox{E}\kern-.125emX}}

\DeclareMathOperator{\sign}{sign}
\DeclareMathOperator{\conv}{Conv1D}
\DeclareMathOperator{\linear}{NLinear}
\DeclareMathOperator{\unif}{Uniform}
\DeclareMathOperator{\init}{Initialize}
\DeclareMathOperator{\nlinear}{NLinear}

\begin{document}

\title{Forecasting Early with Meta Learning}

\author{\IEEEauthorblockN{Shayan Jawed$^*$}
\IEEEauthorblockA{\textit{Information Science and Machine Learning Lab} \\
\textit{University of Hildesheim}\\
Hildesheim, Germany \\
shayan@ismll.uni-hildesheim.de}
\\
\IEEEauthorblockN{Vijaya Krishna Yalavarthi}
\IEEEauthorblockA{\textit{Information Science and Machine Learning Lab} \\
\textit{University of Hildesheim}\\
Hildesheim, Germany \\
yalavarthi@ismll.uni-hildesheim.de}
\and
\IEEEauthorblockN{Kiran Madhusudhanan$^*$}
\IEEEauthorblockA{\textit{Information Science and Machine Learning Lab} \\
\textit{University of Hildesheim}\\
Hildesheim, Germany \\
madhusudhanan@ismll.uni-hildesheim.de}
\\
\IEEEauthorblockN{Lars Schmidt-Thieme}
\IEEEauthorblockA{\textit{Information Science and Machine Learning Lab} \\
\textit{University of Hildesheim}\\
Hildesheim, Germany \\
schmidt-thieme@ismll.uni-hildesheim.de}
}

\maketitle
\def\thefootnote{*}\footnotetext{Equal Contribution}\def\thefootnote{\arabic{footnote}}
\begin{abstract}
In the early observation period of a time series, there might be only a few historic observations available to learn a model. However, in cases where an existing prior set of datasets is available, Meta learning methods can be applicable.
In this paper, we devise a Meta learning method that exploits samples from additional datasets and learns to augment time series through adversarial learning as an auxiliary task for the target dataset. Our model (FEML), is equipped with a shared Convolutional backbone that learns features for varying length inputs from different datasets and has dataset specific heads to forecast for different output lengths. We show that FEML can meta learn across datasets and by additionally learning on adversarial generated samples as auxiliary samples for the target dataset, it can improve the forecasting performance compared to single task learning, and various solutions adapted from Joint learning, Multi-task learning and classic forecasting baselines. 
\end{abstract}

\begin{IEEEkeywords}
Early Time series forecasting, Meta Learning
\end{IEEEkeywords}

\section{Introduction}
\begin{figure}[t]
\centering
\includegraphics[width=0.9\columnwidth]{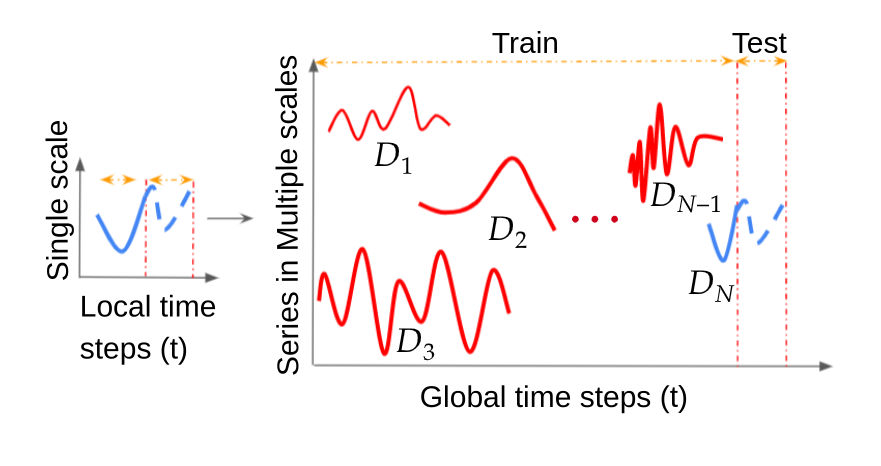}
\caption{Motivation for modeling eTSF as a meta-learning problem. The target eTSF dataset $D_N$ (blue) and the auxiliary datasets $\{D_1,\dots,D_{N-1}\}$ (red). A meta-learning model could utilize the general representation learned across the auxiliary datasets to improve the prediction accuracy on the target dataset where just a few time steps are observed. The dotted line indicates the forecast horizon $h_N$ for target dataset $D_N$.}
\label{fig:introfigure}
\end{figure}

Time series forecasting is an active area of research with wide applications in multiple domains such as Energy resource management \cite{li2019enhancing}, Financial modeling \cite{qin2017dual}, and Environment planning \cite{miau2020river} 
to name a few. For many of these applications to realize, a practical assumption is the availability of Big data \cite{brinkmeyer2022few,iwata2020few}, given most modern research methods rely on Deep learning from vast amounts of data. This motivates the research question of learning accurate forecasting methods from only limited observations. In practice, limited observations could be a result of not observing the time series for a long duration, for example if the frequency of observation is yearly, or due to limited number of data generating processes, for example, availability of only a few sensors. Arguably one of the most impactful applications recently has been forecasting COVID-19 number of cases and deaths where only limited historic observations were available initially. 
Herein, Early Time Series Forecasting (eTSF), the task of forecasting the time series given only limited historic observations, can be modeled. We further motivate the problem setting with Fig. \ref{fig:introfigure}.

However, there is a dearth of research in the direction of general eTSF methods, unlike its classification counterpart of Early time series classification (eTSC) \cite{schafer2020teaser}\cite{gupta2020approaches}.
the early hazard warning forecasting \cite{selva2021probabilistic}, 
or the early manufacturing data forecasting \cite{li2009improved}. Secondly, in comparison to eTSC, eTSF is a more challenging task as it concerns forecasting the full trajectory of the time series from the very limited available data in contrast to solving for in-sample data in classification.

\indent We alleviate the first complication by providing a benchmark for early time series forecasting with 32 datasets collected from the Monash Forecasting Repository\cite{godahewa2021monash} covering various domains. The benchmark could motivate the research community for developing domain-independent models for the task of early forecasting. The second complication translates into an interesting research question, and could be solved by considering eTSF as a direct realization of time series compatible methods from the research areas of Meta learning \cite{brinkmeyer2022few,vaiciukynas2021two}
and Transfer learning \cite{oreshkin2021meta,grazzi2021meta}. 

\indent To the best of our knowledge, meta-learning for the task of eTSF as shown in Figure \ref{fig:introfigure}, is yet to be explored. However, the use of meta-learning methods for general time series forecasting have been explored previously in \cite{oreshkin2021meta,brinkmeyer2022few,grazzi2021meta,jawed2021multi}. 
The pioneering works \cite{oreshkin2021meta,grazzi2021meta} showcased the applicability of well-established forecasting methods to the Zero-shot forecasting problem. The work \cite{brinkmeyer2022few} learned a joint forecasting model across observations from multiple datasets by employing permutation invariant Deep Set blocks \cite{zaheer2017deep} that allow learning with heterogeneous multivariate channels. Among the drawbacks concerning the prior approaches are learning from only a single source dataset \cite{oreshkin2021meta,grazzi2021meta}, learning on a fixed sized input and forecast horizon fixed across datasets\cite{oreshkin2021meta,brinkmeyer2022few}. 
Another limiting factor is the lack of gradient based adaptation \cite{oreshkin2021meta,grazzi2021meta,brinkmeyer2022few} in the scenario of having a larger number of samples (as related time series), a possibility unconstrained by an early time period.

\indent A significant stream of work is dedicated to data augmentation with the similar goal of improving modeling performance in scarce data scenarios. Traditional approaches that inject Gaussian noise or slice windows can lead to a generation of redundant sampling and require extensive hyperparameter tuning for effective coupling with a downstream learning model\cite{wen2020time}. On the other hand, deep generative learning method \cite{yoon2019time},
can augment data in either imbalanced datasets or be applied for anonymization but require substantial data hindering their application for solving the eTSF task.   

\indent In order to solve these challenges, we propose a novel model for \underline{F}orecasting \underline{E}arly with \underline{M}eta \underline{L}earning or \textit{FEML} (pronounced ``fee·mayl'') composed of a convolutional encoder and a stack of linear layer decoders. The convolutional layers learn a joint embedding of input time series across datasets of varying lengths, and the decoding layers enable dataset-specific direct forecasting of the forecast horizons. These Linear layers are further guided by findings in recent studies \cite{zeng2022transformers} that credit the linear layers extrapolating for multiple outputs as the main building block of deep forecasting architectures \cite{jawed2019multi,zhou2021informer,wu2021autoformer,jawed2022gqformer,madhusudhanan2023u}. We further adapt a well-established meta-learning algorithm \textit{Reptile}\cite{nichol2018first} to serialized learning across time series datasets with our multi-head forecasting model FEML. Moreover, we design a novel multi-task loss that enables additional learning on adversarially generated samples through the Fast Gradient Sign Method \cite{goodfellow2014explaining} for the target dataset. This further improves the forecasting performance and the applicability of this augmentation scheme is unhindered by lack of samples and generation is guided by tightly coupled learning on the multi-task loss.

We summarize the contributions as follows:
\begin{enumerate}
    \item To the best of our knowledge, we are the first to address the problem of Early time series forecasting with a principled meta learning solution. 
    \item We design our method to learn across multiple time series datasets, with varying input and forecast lengths.
    \item We provide the first useful benchmark for Early time series forecasting across 32 datasets from the Monash Forecasting Repository \cite{godahewa2021monash}. 
    \item We investigate the effect of data augmentation through adversarial learning.
    \item We benchmark our method against Statistical baselines, Stand-alone State-of-the-art forecasting models and other meta-learning methods like Joint learning and Multi-task learning.    
\end{enumerate}



\section{Related Works}

Statistical methods for time series forecasting positions themselves as strong baselines while forecasting within the low data regime. Methods like ARIMA \cite{godahewa2021monash} and ETS \cite{hyndman2008forecasting,godahewa2021monash} are carefully curated techniques for time series forecasting and have been the default method for explainable forecast from previous decades. ARIMA models are simple regression models where previous lagged observations are used to generate forecast. Simple exponential Smoothing (ETS) computes the forecasts as exponentially decaying weighted averages of past time series observations. 
However, until the introduction of TBATS \cite{de2011forecasting}, none of the statistical baselines were equipped to handle multiple seasonality, high-frequency seasonality, non-integer seasonality and dual calendar effects. TBATS fuses together Box-Cox transformations \cite{box1970time} and Fourier representations along with ARMA error corrections. Since the statistical models leverage only the last few data instances for forecasting the future time steps, they should excel in theory for the task of eTSF. Nevertheless, such a comparison is missing in prior works on meta-learning for forecasting \cite{jin2022domain,brinkmeyer2022few,iwata2020few}.

Modern deep learning architectures based on CNN\cite{borovykh2017conditional}, RNN\cite{rangapuram2018deep,salinas2020deepar} and Transformer \cite{zhou2021informer,li2019enhancing} have also been explored for forecasting. Recently, however, the work from \cite{zeng2022transformers} has showcased that for most time series forecasting tasks, even linear representations can suffice. For the task of long horizon forecasting, linear methods like DLinear and NLinear \cite{zeng2022transformers} were found to outperform counterpart non-linear deep architectures. In essence, the NLinear model is a simple feed-forward network that is meticulously designed to reduce the domain shift while working with long horizon forecasting. Although long horizon forecasting is an interesting topic, it is yet unclear if deep learning methods for time series forecasting would outperform the statistical counterparts for eTSF.

Transferring knowledge from one task to improve performance in a test dataset has been extensively researched within the deep learning community. In Computer vision, models are consistently trained on a large corpus of ImageNet \cite{deng2009imagenet} as a pre-training step before fine-tuning the model for the target task. Within the time series forecasting community, transfer learning was used in \cite{xu2020hybrid} to improve the electricity demand forecast by transferring knowledge from a source location to that of a target location. The authors achieve the objective by decomposing the time series and transferring knowledge from the decomposed components. In \cite{oreshkin2021meta}, the authors propose a zero-shot training strategy that learns on a source dataset and predicts on a target dataset without learning on the target dataset. The authors interpret the zero-shot results as an evidence supporting the use of meta learning for time series forecasting. Even though, the aforementioned studies successfully transfer knowledge across datasets, the transfer of knowledge is limited from one dataset. The work from \cite{iwata2020few} uses a Recurrent neural network equipped with context based Attention for learning meta representations by learning to forecast time series across datasets. In \cite{brinkmeyer2022few} a meta learning model that is able to deal with the problem of having heterogeneous channels was proposed. The method encodes the time series channels with a recurrent network based encoder and then uses Deep Sets \cite{zaheer2017deep} for learning invariant representations across the channel space, effectively providing forecasts for multivariate time series with varying number of channels. Another recent approach that aims to transfer representations \cite{jin2022domain} shares attention representations between a source dataset with many samples and target dataset with fewer samples. Through sharing the attention representations and learning a discriminator that distinguishes samples between the two datasets, the forecasting performance was shown to be improved for the target dataset. In \cite{hu2020datsing} the task of domain adaptation for time series forecasting was explored with augmenting the target dataset with similar samples from various other time series datasets based on computing the DTW based distances \cite{ratanamahatana2004making}.
Among the gradient based approaches to meta-learning for time series, \cite{arango2021multimodal} focuses on learning multi-modal meta representations. The intuition being that a time series can cover multiple modes and thought of as collection of multiple intra-time related tasks to be modeled. Another recent approach in this direction is also noted in \cite{woo2022deeptime}.

We delineate from prior works as firstly, we focus on Early time series forecasting where, to the best of our knowledge, there exists no work and secondly, our work eliminates the restriction set by previous meta-learning techniques to have a fixed forecast horizon and input length across datasets \cite{hu2020datsing,oreshkin2021meta,jin2022domain,iwata2020few,brinkmeyer2022few}. By treating these characteristics as a function of the dataset, we allow for a principled meta-learning solution across datasets. Prior work\cite{hu2020datsing,oreshkin2021meta,jin2022domain} is also limited to learning only from a single auxiliary dataset.

\section{Problem Formulation}

A time series dataset $D$ consists of tuple of time series $D:=\{(X, Y)| (X,Y) \sim \rho\}$ with $X \in \R^\delta$, $Y\in \R^h$, are the predictors and forecasts with $\delta$ being observation time and $h$ being forecast horizon, drawn from a random distribution $\rho$. We denote the domain of the dataset as $\Omega$. We consider a meta dataset $\mathcal{D}$ consists of $N$ many datasets with $D_i \in \mathcal{D}$ with domain $\Omega^i$ drawn from distribution $\rho_i$, with $\delta_i$ and $h_i$ as the observation time range and forecast horizon. For notational convenience, we denote the number of samples per dataset as $M_i$. Further, we denote $D_{N}$ as the support for the target dataset with $\delta_N \ll \delta_i, i<N $. We define the loss function $\ell: \R^{h_{N}} \times \R^{h_{N}} \to \R$. Our objective then is to find a meta-model $m:\Omega^1\times...,\times \Omega^{N-1}\times \R^{\delta_{N}} \to \R^{h_{N}}$, such that the expected loss on $(X,Y) \sim \rho_N$, observed after the occurrence of $D_N$ in chronological order, is minimized:
\begin{align*}
\min \displaystyle \mathop{{}\mathbb{E}}_{(X,Y)\in \rho_N} \ell (Y, m(X, D_1,...,D_{N}))    
\end{align*}
\begin{figure}[t]
\centering
\includegraphics[width=\columnwidth]{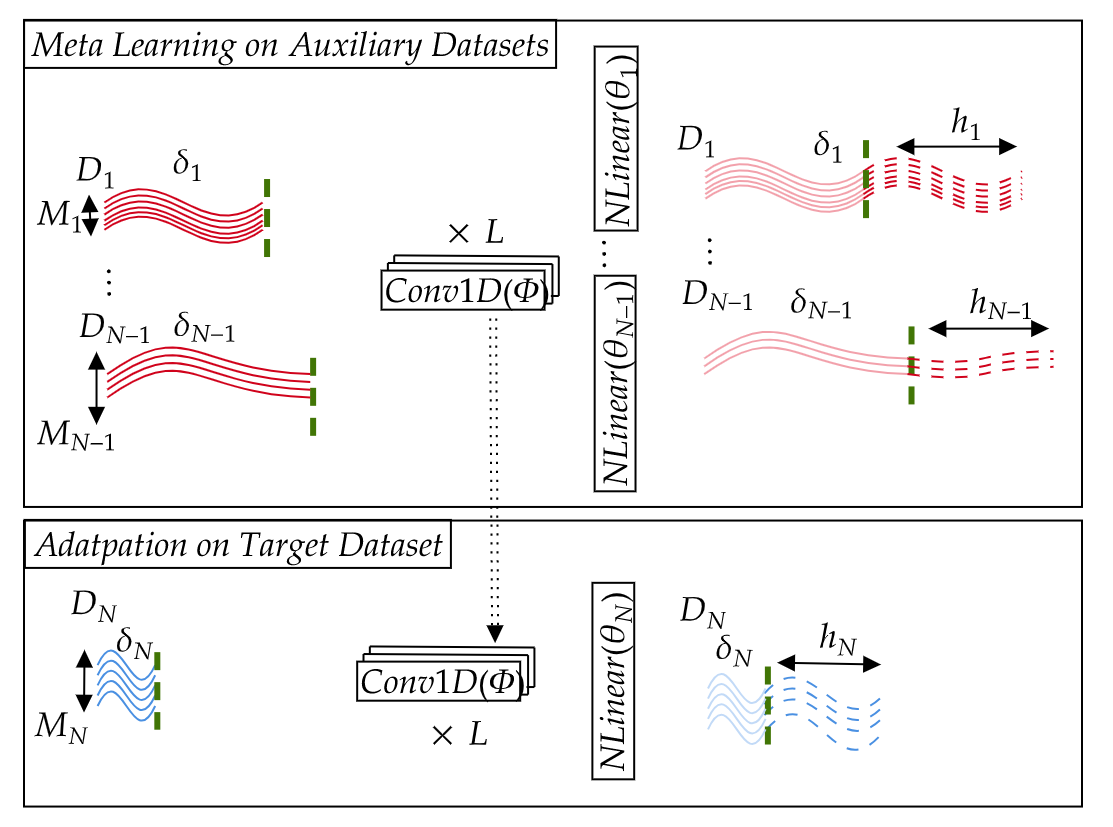}
\caption{\texttt{FEML} learns over the samples from the auxiliary datasets $\{D_1, \dots, D_{N-1}\}$ (represented in red) during the inner iteration and adapts the parameters to the target dataset $D_N$ (represented in blue) in the outer loop. The \texttt{FEML} model consists of shared $\conv(\Phi)$ parameters and dataset specific $\nlinear(\theta_i)$ for $i \in \{1, \dots, N\}$. Here, $\delta, h, M$ indicate the observed range, the forecast horizon and number of tuples in a time series dataset $D$ respectively.}
\end{figure}
\section{Method}
\label{sec:method}
Our forecasting method, \texttt{FEML} learns input length invariant convolutional features for accurate forecasting of multiple datasets. The inputs to \texttt{FEML} are fed through a series of one dimensional convolution layers and non-linear activations. These convolutional features can learn rich locality information. Given that the convolutional features are learned by sliding the 1D kernels over the input window, the convolutional encoding can process input of different lengths from different datasets, expressed as: 
\begin{equation}
    Z_i = \conv(X_i; \Phi)
    \label{eq:convEmb}
\end{equation}
The Encoder $\conv(.;\Phi):\mathbb{R}^{1 \times \delta_i} \to \mathbb{R}^{d \times \delta_i}$ is composed of $L$ many layers and its learnable parameters are jointly denoted as $\Phi$. Passing the input time series from the $i^{th}$ dataset, we get the embedding: $Z_{i} \in \mathbb{R}^{ d \times \delta_i}$. Where $d$ represents the dimensionality of the convolutional filters. 

Importantly, these convolutional features also constitute the shared embedding parameters in \texttt{FEML} ensuring that multiple datasets are embedded in a joint feature space. 
This distinction is important, as later we will see that we perform meta-learning updates on the shared parameters of the network. Furthermore, the use of shared parameters is also guided by a key-insight in Multi-task learning literature which points to the fact that for optimal Multi-task learning a combination of shared, and task-specific parameters is desired. 
Therefore, we design multiple output heads composed of task-specific Linear layers that can forecast for each dataset respectively. These layers take the Convolutional encoding $Z_i$ from the Encoder, and then learn to extrapolate the forecast directly for all forecast outputs. Our choice for stacking task-specific layers besides having a combination of shared and non-shared parameters among tasks is also based on recent findings from \cite{zeng2022transformers} where it was posited that \textit{direct} forecasting with a normalized Linear layer (NLinear), in recent forecasting architectures was key to their success. The output from the $i^{th}$ head can be expressed as:
\begin{equation}
    \hat{Y}_i = \linear_i(Z_i; \theta_i)
    \label{eq:modelPrediction}
\end{equation}
The $\linear_i(.;\theta_i): \mathbb{R}^{d \times \delta_i} \to \mathbb{R}^{h_i}$ where layer parameters $\theta_i$ denote the $i^{th}$ task specific parameters for forecasting a fixed horizon per dataset specified apriori. To further elaborate this, although the $\conv(.;\Phi)$ layers can process inputs of various lengths from different datasets and learn shared parameters for those, the $\linear_i(.;\theta_i)$ layer takes a flattened fixed length vector and outputs for a fixed sized horizon. Therefore, the $\linear_i(.;\theta_i)$ layer is defined for each dataset separately, to forecast for the corresponding forecast horizons. 
\begin{algorithm}[t]
  \caption{Learning for \texttt{FEML}}\label{learning}
  \begin{algorithmic}[1]
  
      \Require{Time series datasets $\{D_1, ..., D_N\}$, learning rates $\left( \mu^{in},\mu^{out},\mu^{ad}, \epsilon, w \right)$, $\Phi$, $\{\theta_i \mid  i=1,...,N\}$}
      \State $\init \; \Phi$
      \State $\init \; \{\theta_i \mid  i=1,...,N-1\}$ 
      
      \While{not converged}
        \State Draw $i \sim \unif(\{1,...,N-1\})$
        \State $\Phi^{in} \gets \Phi$
        \State $\theta_{i}^{in} \gets \theta_{i}$
      \For{ $\left( X_i, Y_i \right) \in D_i$}
        \State Compute $\hat{Y_i}$ \Comment{using Eq. \ref{eq:convEmb},\ref{eq:modelPrediction}}
        \State Update $\Phi^{in}, \theta_i^{in}$ \Comment{using Eq. \ref{eq:innerupdate}}
    \EndFor
      \State Update meta parameters $\Phi, \theta_i$ \Comment{using Eq. \ref{eq:metaupdate}}

      \State $\Phi^{ad} \gets \Phi$
      
      \State $\init \theta_N$

      \For{$\left(X_N, Y_N\right) \in D_N$}
        \State Generate $X'_N$\Comment{using Eq.~\ref{eq:adversarialSample}}
        \State Compute $\hat{Y}_N, \hat{Y}'_N$ for $X_N, X'_N$ \Comment{using Eq. \ref{eq:convEmb},\ref{eq:modelPrediction}}
        \State Compute $\ell_N$ \Comment{using Eq. \ref{eq:mtlLoss}}
      \State Update $\Phi^{ad}, \theta_N$ \Comment{using Eq. \ref{eq:adupdate}}
      \EndFor
    
\EndWhile\label{euclidendwhile}
   \State Output $\Phi^{ad}$, $\theta_N$
\end{algorithmic}
\label{al:algorithm}
\end{algorithm}
This construction resembles architectures with multiple heads to solve multi-task problems in computer vision tasks \cite{misra2016cross, kendall2018multi, liu2019end}. We can also contrast this construction to a single head decoding model as presented in the ablation. 

\subsection{Adversarial learning with \texttt{FEML}}
Adversarial samples are samples generated from a similar input data distribution that could confuse the model to output wrong predictions. We follow a well-established approach, Fast Gradient Sign Method (FGSM) \cite{goodfellow2014explaining} to generate adversarial samples. Given input $\left( X_{N}, Y_{N}\right) \in D_N$ the FGSM method computes an adversarial perturbation by taking a gradient step in the loss maximization direction with respect to the input. This can be expressed simply as:
\begin{equation}
    X_N' = X_N + \epsilon \cdot \sign\left(\frac{\partial}{\partial X_N} l \left(Y_N, \hat{Y}_N \right)\right)
    \label{eq:adversarialSample}
\end{equation}
Where $\epsilon$ is the learning rate, or the amount of perturbation based on the sign of the gradient that is added to the input to maximize the loss. Note that we only generate the adversarial samples for the target dataset samples, $(X_N, Y_N)$. Our task then is to incorporate these adversarial samples as additional training samples for the meta-model $m(.)$, effectively doing data augmentation for the target dataset. Consequently, we can generate $\hat{Y}'_N$ as the model forecasts for the adversarial samples $X'_N$ following similar Eq. \ref{eq:convEmb},\ref{eq:modelPrediction}. This leads to a multi-task loss formulation combining learning on the augmented and the original samples using a weighting parameter $w$ to increase model robustness:
\begin{equation}
    \ell_{N} = \ell(Y_N, \hat{Y}_N) + w \cdot \ell(Y_N, \hat{Y}'_N) 
    \label{eq:mtlLoss}
\end{equation}
\subsection{Meta learning with \texttt{FEML}}
\label{sec:MetaL}
We aim to learn an initialization $\Phi$ from a collection of tasks $i \in \{1,...,N-1\}$ such that, in only a few optimization steps, the parameters $\Phi$ can be adapted to solving a new task $N$. We adapt a serialized version of \texttt{Reptile}\cite{nichol2018first} to learn across datasets with \texttt{FEML}. We initialize, the two sets of parameters, $\Phi^{in} \gets \Phi$, $\theta^{in}_{i} \gets \theta_i$ and then for a select number of inner epochs, for the $i^{th}$ sampled task, the shared parameters, $\Phi$ and the task specific parameters, $\theta_i$ are updated as follows:
\begin{align}
	\Phi^{in} &\gets \Phi^{in} -  \mu^{in} \frac{\partial}{\partial\Phi^{in}} \ell \left( Y_i, \hat{Y}_i\right)  \nonumber\\  
        \theta^{in}_i &\gets \theta^{in}_i - \mu^{in} \frac{\partial}{\partial \theta^{in}_i} \ell \left(Y_i,  \hat{Y}_i\right) 
        \label{eq:innerupdate}
\end{align}
The meta-update can be expressed with respect to the meta-gradient, being the difference of the previous and the updated parameters with learning on $i^{th}$ task :
\begin{align}
	 \Phi  &\gets  \Phi- \mu^{out}  ( \Phi - \Phi^{in}) \nonumber\\  
       \theta_{i} &\gets \theta_i - \mu^{out}  ( \theta_i - \theta^{in}_{i} )
       \label{eq:metaupdate}
\end{align}
where ${\mu^{in}, \mu^{out}}$ denote the inner optimization and outer optimization learning rates respectively. In the serialized optimization, after the meta-gradient updates, the meta-parameters are adapted to the target dataset. This involves learning on few samples from the target dataset, $(X_N, Y_N) \in D_N$. 

For the target dataset, a new head is initialized, $\theta_N$. We start the adaptation by copying the shared parameters $\Phi$ into another set $\Phi^{ad}$. The parameters updates can be expressed similar to before, however, with the adversarial multi-task loss, $\ell_N$: 
\begin{align}
	\Phi^{ad} &\gets \Phi^{ad} - \mu^{ad} \frac{\partial}{\partial\Phi^{ad}} \ell_N \left( Y_N, \hat{Y}_N, \hat{Y}'_N  \right)\nonumber\\  
        \theta_N &\gets \theta_N - \mu^{ad} \frac{\partial}{\partial \theta_N} \ell_N \left(Y_N, \hat{Y}_N, \hat{Y}'_N \right) 
        \label{eq:adupdate}
\end{align}
The optimization routine then involves sampling another task $i \in [1,...,N-1]$ and updating the parameters by reinitializing $\Phi^{in} \gets \Phi$, $\theta^{in}_{i} \gets \theta_i$ with the meta-gradient updates before the adaptation. The serial adaptation thus ensures that the model parameters are not overfit on the target dataset and only adapted through one epoch over the target dataset samples after learning on the $i^{th}$ randomly sampled task.

We can note that the above optimization steps involve updating the shared parameters and only the parameters from the Linear layer corresponding to the task sampled. This is in contrast to the standard update in \texttt{Reptile} where no such differentiation is made with regard to task specific and shared parameters, and all parameters are updated jointly. We summarize the method in Algorithm \ref{al:algorithm}.


\section{Experiments}

\subsection{Datasets}
\label{sec:datasets}

We base our experiments on the first comprehensive time series forecasting repository composed of 25 publicly available time series datasets published by \cite{godahewa2021monash}. The datasets can be differentiated by distinctive forecast horizons, series lengths, seasonalities and missing values. Further aggregation across multiple frequencies like monthly, yearly etc., leads to 43 available time series datasets in total.

In our experiments, we limit ourselves to work with 32 datasets from the available 43, by ignoring datasets without date-time information and the datasets with a single univariate series. Additionally, the Monash archive provides the forecast horizon $h_i$ and the lag (observation range) $\delta_i$ defined per dataset $D_i$ based on the human expert evaluation for these datasets. As a heuristic, the lag $\delta_i$ is computed as the seasonality multiplied with a factor of 1.25. By using this large collection of datasets, we can study the performance of the benchmarked baselines and proposed model comprehensively. We use Intel(R) Xeon(R) CPU E5645 for running all statistical baseline experiments and NVIDIA GeForce GTX 1080 Ti for the deep learning experiments. We mark  baselines that required extensive run time (more than 5 days) using “-” symbol within the results table. All reported results are the mean of 3 runs. The code\footnote{https://github.com/super-shayan/FEML} is open sourced for reproducibility. 


\subsection{Evaluation Protocol}
\label{sec:exp}
 \subsubsection{Validation}
A trivial choice for validation is to use out-of-sample data \cite{zeng2022transformers, zhou2021informer}. However, within the low data regime, in-sample strategy becomes a more suitable option \cite{salinas2020deepar}.
For each dataset $D_i$, we randomly choose $10\%$ percent of the available $M_i$ many univariate time series for in-sample validation and train on the rest $90\%$ of the data. In case of datasets with fewer than $10$ time series samples, we randomly assign $1$ of these time series as validation, restricting us to work with datasets with more than $1$ series.

\subsubsection{Test}
During testing, the model forecasts for the next forecast horizon that was unseen during training. 

\subsection{Baselines}

\textbf{Naive Baselines}: 
Our battery of experiments includes benchmarking the following Naive baselines.
\begin{enumerate}
    \item \textit{Mean Forecast}: Repeats the mean of the input as the forecast for the entire horizon.
    \item \textit{Naive Forecast}: Copies the last season as the forecast.
\end{enumerate}

\textbf{Statistical Baselines} : 
In order to judge the performance of our method, we compare against well established statistical baselines:
\begin{enumerate}
    \setcounter{enumi}{2}
    \item \textit{ARIMA} \cite{box1970time}: 
    We fit each univariate time series by varying lagging and differencing hyperparameters. 
    \item \textit{ETS} \cite{hyndman2008forecasting}: 
    ETS fits the best exponential smoothing parameters automatically for a given time series.
    \item \textit{TBATS} \cite{de2011forecasting}: TBATS handles multiple seasonal periods within a time series. 
    TBATS automatically fits the time series by selecting the best seasonality parameters, transformation parameters etc.
\end{enumerate}

\textbf{Single task Learning Baselines}:
Another interesting set of baselines are the boosting models and current state-of-the-art time series forecasting models. 
\begin{enumerate}
    \setcounter{enumi}{5}
    \item XGBoost\cite{chen2016xgboost}: Gradient boosted methods are ensemble models of decision trees. We tune the number of estimators on a grid of $[100, 200, 400, 800]$ per dataset \cite{elsayed2021we}.
    
    \item NLinear \cite{zeng2022transformers}: 
    NLinear is a normalized single feed-forward network and requires no hyperparameter tuning.
    
\end{enumerate}
\subsection{Evaluation}
All models are trained using Mean Absolute Error (MAE) \cite{godahewa2021monash}. 
Table \ref{tb:maintable} reports results across multiple datasets, using the MASE metric. For $M_N$ samples, the MASE is computed by normalizing the MAE of the evaluated model forecast $\hat{Y}$ with the MAE 
 corresponding to the Naive forecast $\hat{Y}^{\text{naive}}$: 
\begin{align}
    \text{MASE}&=\frac{1}{M_N}\sum_{j=0}^{M_N} \frac{|Y_j - \hat{Y}_j|}{|Y_j - \hat{Y}^{\text{Naive}}_j|}
\end{align}
 It avoids symmetry issues, scaling issues and division by $0$ issues and is a generally applicable scale free measure to compare across datasets \cite{godahewa2021monash}. We use \emph{leave-one-out} strategy during evaluation, by considering one dataset as target dataset $D_N$ for eTSF and assuming that the rest of the datasets $D_i \in \{D_1, \dots, D_{N-1}\}$ are observed before $D_N$.

\section{Results}

Table \ref{tb:maintable}, summarizes the experimental results comparing the proposed \texttt{FEML} model with single task baseline on 32 datasets with varying (1) number of time series samples per dataset $M_N$, (2) forecast horizon $h_N$ and (3) lag $\delta_N$. We sort the table in the increasing order of samples for comparative purposes. Here, we try to answer the following research questions:

\indent \textbf{RQ1}: How does the proposed \texttt{FEML} compare against \indent classical and single-task methods?  \\
\indent \textbf{RQ2}: Which statistical baselines are better suited for the \indent task of eTSF? \\
\indent \textbf{RQ3}: How do the statistical baselines fair against their \indent counterpart deep learning baselines?

\subsection*{\textbf{RQ1}: \texttt{FEML} vs classical and single-task methods}

From Table \ref{tb:maintable}, it is clear that the proposed \texttt{FEML} model outperforms the baselines by winning in 12 out of the 32 or 37.5\% of all the datasets. The ability of the proposed \texttt{FEML} model to distil useful information across datasets along with adversarial augmentation technique evidently provides the model an advantage for the task of eTSF. 
For context, the reported results in Monash forecasting archive \cite{godahewa2021monash} show that the best performing method is able to win in 19\% of datasets for the task of general time series forecasting. Transforming the general time series forecasting  dataset to eTSF adds another complexity that only a few observations from target datasets are available. Moreover, the proposed \texttt{FEML} method wins in twice as many datasets versus the second-best baseline.

\subsection*{\textbf{RQ2}: Comparison of statistical baselines for eTSF}
As described before, in the eTSF setting, statistical methods should be viewed as strong baselines. Among the statistical baselines, ARIMA model outperforms other statistical baselines, especially for short forecast horizons. In general, TBATS, being a more sophisticated technique and containing ARMA errors, should outperform the simpler ARIMA model, however, tuning the large hyperparameter space in TBATS might restrict the performance of the model in low data regime.

\begin{table}[t]
\caption{Comparison of the proposed \texttt{FEML} with single task baseline methods based on mean MASE metric. $M_N, \delta_N, h_N$ indicate the number of univariate series, input sequence length and the forecast horizon for the target dataset respectively.}
\centering
\label{tb:maintable}
\setlength{\tabcolsep}{2pt}
\resizebox{\columnwidth}{!}{%
\begin{tabular}{lcccccccccc}
\hline
\textbf{Dataset}   & \textbf{$M_N$} & \textbf{$\delta_N$} & \textbf{$h_N$} & \textbf{Mean}                          & \textbf{Arima}                         & \textbf{ETS}                           & \textbf{TBATS}                         & \textbf{XGBoost} & \textbf{NLinear}                       & \textbf{FEML}                          \\ \hline
Aus. Elecdemand    & 5          & 420                            & 336                          & 2.009                                  & 2.045                                  & 2.164                                  & 1.195                                  & 1.574            & 1.188                                  & \cellcolor[HTML]{9FC5E8}\textbf{0.969} \\
Bitcoin            & 18         & 9                              & 30                           & 0.989                                  & \cellcolor[HTML]{9FC5E8}\textbf{0.684} & 0.846                                  & 1.026                                  & 1.391            & 0.889                                  & 1.025                                  \\
Pedestrians        & 66         & 210                            & 24                           & 1.514                                  & 1.501                                  & 2.466                                  & 0.897                                  & 0.493            & 0.546                                  & \cellcolor[HTML]{9FC5E8}\textbf{0.475} \\
FRED-MD            & 107        & 15                             & 12                           & 0.574                                  & \cellcolor[HTML]{9FC5E8}\textbf{0.415} & 1.161                                  & 2.318                                  & 1.021            & 1.698                                  & 0.812                                  \\
NN5 Weekly         & 111        & 65                             & 8                            & 0.852                                  & \cellcolor[HTML]{9FC5E8}\textbf{0.691} & 0.720                                  & 0.709                                  & 0.699            & 1.179                                  & 0.734                                  \\
NN5 Daily          & 111        & 9                              & 56                           & 0.900                                  & 0.874                                  & 0.883                                  & 0.941                                  & 0.645            & \cellcolor[HTML]{9FC5E8}\textbf{0.587} & 0.596                                  \\
Solar Weekly       & 137        & 6                              & 5                            & 1.496                                  & \cellcolor[HTML]{9FC5E8}\textbf{0.554} & 1.277                                  & 1.189                                  & 1.368            & 1.681                                  & 1.366                                  \\
M1 Yearly          & 179        & 2                              & 6                            & 1.020                                  & 0.920                                  & 1.006                                  & 1.253                                  & 1.337            & 0.994                                  & \cellcolor[HTML]{9FC5E8}\textbf{0.639} \\
M1 Quarterly       & 203        & 5                              & 8                            & 0.933                                  & 0.944                                  & 0.992                                  & \cellcolor[HTML]{9FC5E8}\textbf{0.930} & 0.940            & 1.152                                  & 1.059                                  \\
COVID              & 266        & 9                              & 30                           & 1.007                                  & \cellcolor[HTML]{9FC5E8}\textbf{0.948} & 0.976                                  & 0.952                                  & 1.213            & 0.950                                  & 0.974                                  \\
KDD                & 270        & 210                            & 168                          & \cellcolor[HTML]{9FC5E8}\textbf{0.672} & 0.947                                  & 1.037                                  & 0.997                                  & 0.714            & 0.863                                  & 0.791                                  \\
Electricity Weekly & 321        & 65                             & 8                            & 1.409                                  & 1.361                                  & 1.767                                  & \cellcolor[HTML]{9FC5E8}\textbf{0.884} & 2.882            & 0.902                                  & 1.267                                  \\
Electricity Hourly & 321        & 30                             & 168                          & 2.472                                  & 2.360                                  & 3.648                                  & \cellcolor[HTML]{9FC5E8}\textbf{1.037} & 2.471            & 1.325                                  & 1.313                                  \\
Vehicle Trips      & 329        & 9                              & 30                           & 0.954                                  & 1.020                                  & 1.015                                  & 1.069                                  & 1.133            & \cellcolor[HTML]{9FC5E8}\textbf{0.792} & 0.882                                  \\
M4 Weekly          & 359        & 65                             & 13                           & 0.999                                  & 0.845                                  & 0.824                                  & \cellcolor[HTML]{9FC5E8}\textbf{0.728} & 0.837            & 0.891                                  & 0.776                                  \\
Tourism Monthly    & 366        & 15                             & 24                           & 1.106                                  & 1.075                                  & 1.122                                  & 0.988                                  & 0.817            & \cellcolor[HTML]{9FC5E8}\textbf{0.624} & 0.625                                  \\
M4 Hourly          & 414        & 210                            & 48                           & 2.224                                  & 1.393                                  & 2.353                                  & 0.805                                  & 1.029            & 0.992                                  & \cellcolor[HTML]{9FC5E8}\textbf{0.730} \\
Tourism Quarterly  & 427        & 5                              & 8                            & 0.819                                  & 0.869                                  & 0.754                                  & 0.884                                  & 0.750            & \cellcolor[HTML]{9FC5E8}\textbf{0.537} & 0.621                                  \\
Tourism Yearly     & 518        & 2                              & 4                            & 0.973                                  & 0.949                                  & 0.928                                  & 0.842                                  & 1.017            & 0.951                                  & \cellcolor[HTML]{9FC5E8}\textbf{0.633} \\
M1 Monthly         & 617        & 15                             & 18                           & 0.917                                  & 0.916                                  & 0.874                                  & 1.003                                  & 1.174            & \cellcolor[HTML]{9FC5E8}\textbf{0.843} & 0.880                                  \\
M3 Yearly          & 645        & 2                              & 6                            & 1.011                                  & \cellcolor[HTML]{9FC5E8}\textbf{0.948} & 0.998                                  & 1.110                                  & 1.012            & 1.001                                  & 1.304                                  \\
M3 Quarterly       & 756        & 5                              & 8                            & 0.890                                  & 0.882                                  & 0.843                                  & 0.864                                  & 0.795            & 0.809                                  & \cellcolor[HTML]{9FC5E8}\textbf{0.712} \\
Hospital           & 767        & 15                             & 12                           & 0.848                                  & 0.857                                  & 0.910                                  & 0.910                                  & 1.343            & \cellcolor[HTML]{9FC5E8}\textbf{0.752} & 0.841                                  \\
Traffic Weekly     & 862        & 65                             & 8                            & 0.797                                  & 0.560                                  & \cellcolor[HTML]{9FC5E8}\textbf{0.544} & 0.655                                  & 0.716            & 0.620                                  & 0.664                                  \\
Traffic Hourly     & 862        & 30                             & 168                          & 1.420                                  & 1.405                                  & 1.575                                  & 1.051                                  & 1.046            & 1.248                                  & \cellcolor[HTML]{9FC5E8}\textbf{1.045} \\
M3 Monthly         & 1428       & 15                             & 18                           & 0.896                                  & 0.910                                  & 0.873                                  & \cellcolor[HTML]{9FC5E8}\textbf{0.804} & 0.892            & 0.804                                  & 0.896                                  \\
Carparts           & 2674       & 15                             & 12                           & 0.910                                  & 0.982                                  & 0.948                                  & 1.034                                  & 0.908            & 0.899                                  & \cellcolor[HTML]{9FC5E8}\textbf{0.648} \\
M4 Daily           & 4225       & 9                              & 14                           & 0.948                                  & 0.954                                  & 0.801                                  & 0.892                                  & 0.917            & 0.800                                  & \cellcolor[HTML]{9FC5E8}\textbf{0.796} \\
M4 Quarterly       & 23792      & 5                              & 8                            & 0.899                                  & 0.886                                  & 0.821                                  &  -                                      & 0.784            & 0.787                                  & \cellcolor[HTML]{9FC5E8}\textbf{0.767} \\
Temp. Rain         & 32072      & 9                              & 30                           & \cellcolor[HTML]{9FC5E8}\textbf{0.861} & 0.904                                  & 0.945                                  & 1.521                                  & 1.189            & 0.904                                  & 0.941                                  \\
M4 Monthly         & 47776      & 15                             & 18                           & 0.928                                  & 0.783                                  & 0.704                                  &   -                                     & 0.727            & 0.707                                  & \cellcolor[HTML]{9FC5E8}\textbf{0.683} \\
Kaggle Weekly      & 145063     & 10                             & 8                            & 0.977                                  & 0.963                                  & 0.991                                  &  -                                      & 1.113            & 0.941                                  & \cellcolor[HTML]{9FC5E8}\textbf{0.879} \\ \hline
\multicolumn{4}{r}{Wins}                                                                        & 2                                      & 6                                      & 1                                      & 5                                      & 0                & 6                                      & 12                                     \\ \hline 
\multicolumn{4}{r}{$\%$ Wins}                                                            & 6.25                                   & 18.75                                  & 3.125                                  & 15.625                                 & 0                & 18.75                                  & 37.5     \\\hline                             
\end{tabular}%
}
\end{table}

\subsection*{\textbf{RQ3}: Statistical baselines vs Learned baselines}
Among the learned baselines, NLinear performs at-par with the ARIMA model by winning in 6 out of the 32 datasets, whereas, the XGBoost model is unable to win in any of 32 datasets, even though the complexity for the XGBoost model was tuned by varying the number of estimators.

\begin{figure*}[h]
    \rotatebox{90}{\makebox[0pt][c]{\quad \quad \quad \quad \quad \quad \quad \quad \quad \quad \quad \quad \scriptsize Wins}}
  \centering
  \subfloat[RQ4: Multi Horizon Comparison]{\includegraphics[width=.32\textwidth]{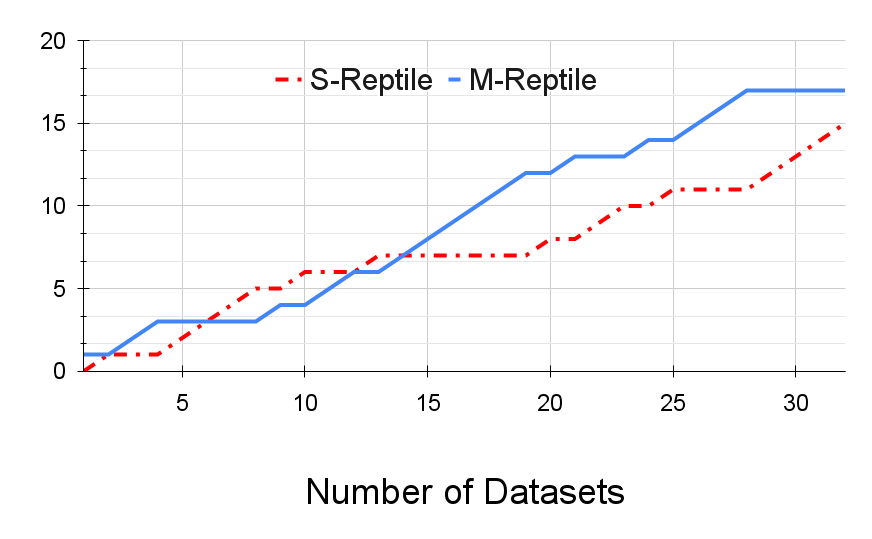}\label{fg:RQ1}}
  \subfloat[RQ5: Adversarial NLinear]{\includegraphics[width=.32\textwidth]{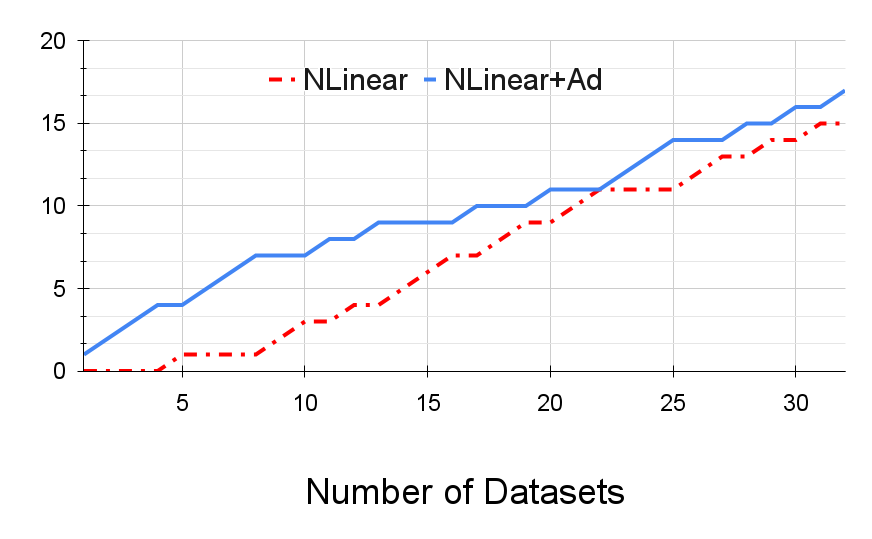}\label{fg:RQ2nlin}}
  \subfloat[RQ6: Adversarial M-Reptile]{\includegraphics[width=.32\textwidth]{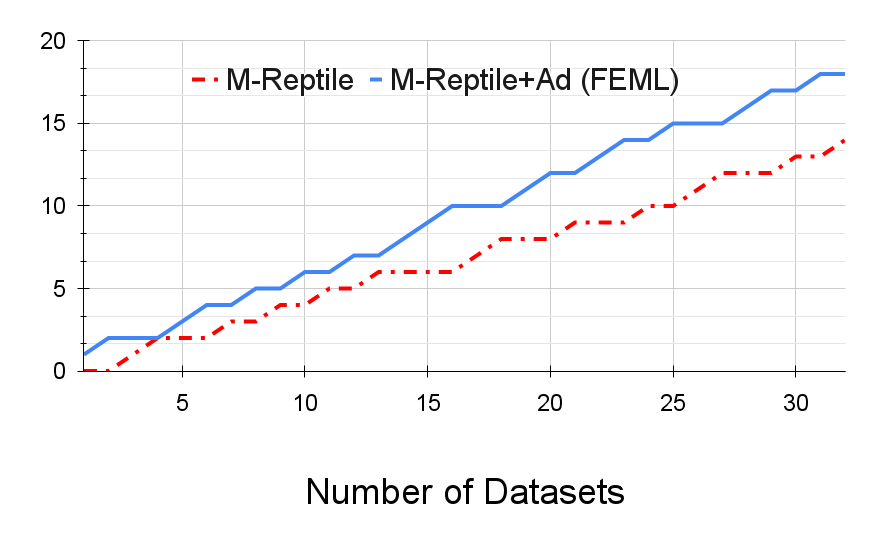}\label{fg:RQ2rept}} \\
  \vspace{0.2cm}
  \vspace{0.2cm}
  \caption{X-axis represents the number of datasets and the Y-axis represents the wins gathered across the datasets. (a) Comparison of wins across datasets for Multi-Head Reptile model (M-reptile) and shared Single-Head Reptile model (S-Reptile). (b) Plot showing the increase in wins while using adversarial augmentation along with base NLinear base model. (c) Plot showing the increase in wins while using adversarial augmentation along with M-Reptile base model (also proposed as \texttt{FEML}).}
  \label{fig:sub1}
\end{figure*}


\section{Ablation}

In this section, we try to analyze each component of the proposed \texttt{FEML} model for eTSF. 

\indent \textbf{RQ4}: Does the proposed multi-head architecture perform \indent better than a single shared head architecture?

\indent \textbf{RQ5}: Does the proposed adversarial augmentation strategy \indent improve performance of a  base model, for eTSF?

\indent \textbf{RQ6}: How does the proposed \texttt{FEML} compare against \indent other meta-learning methods for eTSF?


The experiments were conducted on the same 32 benchmark datasets and using a similar experimental setup as described in Section \ref{sec:exp}. We present the results of the experiments as the number of wins in a head-to-head comparison Fig. \ref{fig:sub1}, where the x-axis shows the datasets arranged in ascending order of the number of samples per dataset ($M_N$), and the y-axis shows the sum of wins across the number of datasets.

\subsection*{\textbf{RQ4} : Multi-head Forecasting Model}

Useful forecast horizon for a particular time series dataset depends on the use-case for that dataset. 
We build our model on top of this premise and compare \texttt{FEML} without adversarial learning (M-Reptile) with that of a single-head model (S-Reptile) to handle the flexible horizons across datasets. 
The S-reptile model is a single head model with a shared last linear layer having as many output neurons as the largest forecast horizon. For datasets requiring fewer forecast horizon, we ignore the outputs produced from the extra output nodes.


Figure \ref{fg:RQ1}, compares the number of wins (y-axis) across the number of datasets (x-axis) for the M-Reptile and S-Reptile model. The results show that M-Reptile model wins on more number of datasets in comparison to the S-Reptile model, indicating that a shared output layer across datasets has an adverse effect on the expressiveness of the model. The M-Reptile model with dataset specific multitask heads helps the model to learn dataset specific parameters, which aids the model in outperforming its shared single head counterpart. 

\subsection*{\textbf{RQ5}: Advantage of Adversarial Augmentation}
In order to understand the advantage offered by the adversarial augmentation, we perform experiments on two baseline models (NLinear and M-Reptile) and show that the adversarial augmentation improves the number of wins in comparison to the base model. Figure \ref{fg:RQ2nlin} highlights the improvement for a single task baseline NLinear with the addition of adversarial augmented examples. For the single task baseline, adversarial examples prove advantageous, especially when the number of samples within the dataset are few. The adversarial samples improve robustness of the baseline models by providing additional samples of the same distribution as the input data, that mitigates overfitting. Furthermore, the improvement is not limited to single task base models. We performed a similar experiment with the M-Reptile model to interpret the improvement brought about by the adversarial samples on top of a meta-learning baseline. Interestingly, Figure \ref{fg:RQ2rept} shows that the number of wins increases even further while comparing the M-Reptile model with and without adversarial samples. The \texttt{FEML} M-Reptile+Ad model has 4 more wins over the 32 datasets (as can be viewed on the right corner of the image) in a head-to-head comparison with the M-Reptile model. The results show that the adversarial augmentation strategy could be beneficial even in a meta-learning setting. 

\subsection*{\textbf{RQ6}: Comparison to Other Meta-Learning Methods}

We experiment multiple meta-learning methods to determine a useful strategy for eTSF. We introduce three additional baselines for these experiments. All the baselines have similar architecture as the proposed multi-head architecture from \texttt{FEML} and differ only in the learning technique employed.

\begin{enumerate}
    \item \textit{Joint Learning}: The model is trained on a concatenation of all the available source datasets $\{D_1,...,D_{N-1}\}$ and the support samples from the target dataset represented as $D_N$. The model does not learn any task specific embeddings.
    \item \textit{Multitask Learning (MTL)}: 
    In MTL the tasks are segregated as target task and auxiliary tasks. For eTSF, the target task is set as learning on the support samples from the target data represented as $D_N$ and the auxiliary targets are the information accumulated from the auxiliary data $\{D_1,...,D_{N-1}\}$. Mostly, the target task is heavily weighted in comparison to the auxiliary tasks, and we follow the same approach. We set the target task to half of the total weight and share the rest of the weights equally among the auxiliary tasks.
    \item \textit{Reptile}: 
    We employ the Reptile training strategy in \texttt{FEML} and is described in Section \ref{sec:MetaL}.
\end{enumerate}

\begin{figure}[th]
\rotatebox{90}{\makebox[0pt][c]{\quad \quad \quad \quad \quad \quad \quad \quad \quad \quad \quad \quad \quad \quad \quad \small Wins}}
\centering
\includegraphics[width=0.8\columnwidth]{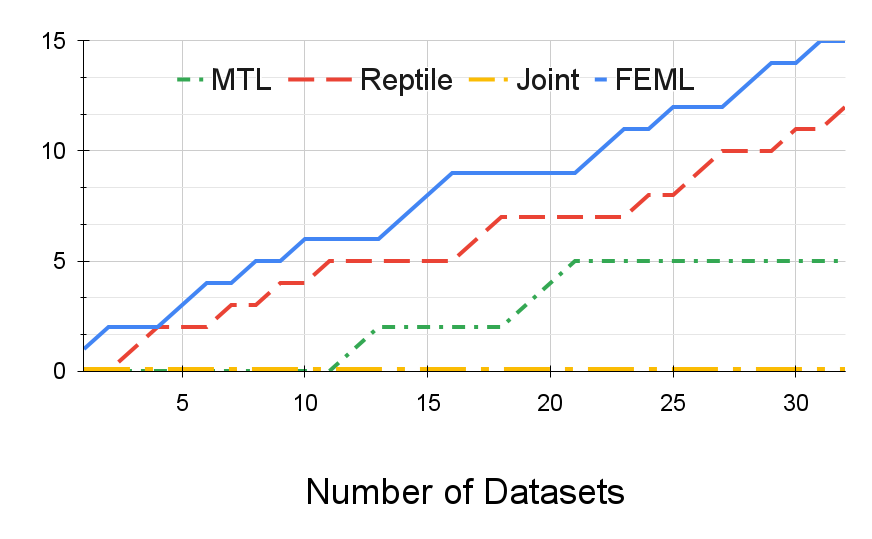}
\caption{Comparison of wins across datasets for various meta-learning baselines, namely Joint Learning (Joint), Multi-task Learning (MTL), Reptile, and the proposed \texttt{FEML}.}
\label{fg:meta}
\end{figure}

The results of the experiment are presented in Figure \ref{fg:meta}. The proposed \texttt{FEML} achieves the highest number of wins when compared to base meta-learning baselines. This could be attributed to the benefit \texttt{FEML} receives from the additional adversarial samples. In comparison to the base MTL and Joint training, Reptile learning algorithm is able to adapt better to the fewer samples available for eTSF. Joint training is unable to win in any of the datasets, and this could be explained away as joint training does not prioritize the target dataset $D_N$ any more than the auxiliary datasets, leading to poor performance overall. Multitask Learning (MTL) is able to win in 5 datasets, however is not as effective as the quick adaptation offered by the reptile learning algorithm. 
\section{Conclusion}

In this paper, we formally define the task of Early Time Series Forecasting (eTSF), which deals with forecasting when very few observations are available from the target time series dataset. We propose the \texttt{FEML} model, a meta-learning based method to assimilate information from other related or unrelated time series datasets to improve performance on the target dataset. In addition, \texttt{FEML} is equipped with the proposed adversarial augmentation strategy that allows \texttt{FEML} to learn from additional augmented examples of the same distribution as the target dataset. Experimental results on 32 real world time series datasets indicate that the proposed \texttt{FEML} model outperforms statistical, single task and other meta-learning baselines. We believe this paper provides a foundation for future work in this direction. 

In future works, we plan to further explore useful augmentation strategies especially for eTSF as learning useful augmented samples from very few observations could prove beneficial. Research in the direction of a hybrid model that carefully integrates statistical models and deep learning or meta-learning models is also an interesting research direction. 

\section{Acknowledgment}

This work was supported by the Federal Ministry for Economic Affairs and Climate Action (BMWK), Germany, within the framework of the IIP-Ecosphere project (project number: 01MK20006D) and the industry partner “VWFS DARC: Volkswagen Financial Services Data Analytics Research Center".
\bibliography{biblio}
\bibliographystyle{IEEEtran}
\end{document}